\pdfoutput=1

\documentclass[11pt]{article}

\usepackage{acl}

\usepackage{times}
\usepackage{latexsym}

\usepackage[T1]{fontenc}

\usepackage[utf8]{inputenc}

\usepackage{microtype}

\usepackage{booktabs}
\usepackage{multirow}
\usepackage{graphicx}
\usepackage{inconsolata}
\usepackage{dirtytalk}
\usepackage{amsmath}
\usepackage{bbding}
\usepackage{amssymb}
\usepackage{pifont}
\newcommand{\cmark}{\ding{51}}
\newcommand{\xmark}{\ding{55}}

\definecolor{MyOrange}{RGB}{230, 159, 0}
\definecolor{MyBlue}{RGB}{86, 180, 233}
\definecolor{MyGreen}{RGB}{0, 158, 115}
\definecolor{MyPurple}{RGB}{204, 121, 167}

\title{A sequence-to-sequence approach for document-level relation extraction}

\author{
  \(\text{John~Giorgi}^{1,4,5,\text{\Envelope}}\;\;
  \text{Gary~D.~Bader}^{1,2,4,6,7,\dag}\;\; \text{Bo~Wang}^{1,3,5,8,\dag}\;\;\) \\
  \(^{1} \text{Department of Computer Science, University of Toronto}\)\\
  \(^{2} \text{Department of Molecular Genetics, University of Toronto}\) \\
  \(^{3} \text{Department of Laboratory Medicine and Pathobiology, University of Toronto}\) \\
  \(^{4} \text{Terrence Donnelly Centre for Cellular \& Biomolecular Research}\) \\
  \(^{5} \text{Vector Institute for Artificial Intelligence}\) \\
  \(^{6} \text{The Lunenfeld-Tanenbaum Research Institute, Sinai Health System}\) \\
  \(^{7} \text{Princess Margaret Cancer Centre, University Health Network}\) \\
  \(^{8} \text{Peter Munk Cardiac Center, University Health Network}\) \\
  \(^{\text{\Envelope}} \text{Corresponding author} \;\; ^{\dag} \text{Equal contribution}\) \\
  \texttt{\{john.giorgi, gary.bader\}@mail.utoronto.ca} \\
  \texttt{bowang@vectorinstitute.ai}\\}

\begin{document}
\maketitle
\begin{abstract}
Motivated by the fact that many relations cross the sentence boundary, there has been increasing interest in document-level relation extraction (DocRE). DocRE requires integrating information within and across sentences, capturing complex interactions between mentions of entities. Most existing methods are pipeline-based, requiring entities as input. However, jointly learning to extract entities and relations can improve performance and be more efficient due to shared parameters and training steps. In this paper, we develop a sequence-to-sequence approach, seq2rel, that can learn the subtasks of DocRE (entity extraction, coreference resolution and relation extraction) end-to-end, replacing a pipeline of task-specific components. Using a simple strategy we call entity hinting, we compare our approach to existing pipeline-based methods on several popular biomedical datasets, in some cases exceeding their performance. We also report the first end-to-end results on these datasets for future comparison. Finally, we demonstrate that, under our model, an end-to-end approach outperforms a pipeline-based approach. Our code, data and trained models are available at {\small{\url{https://github.com/johngiorgi/seq2rel}}}. An online demo is available at {\small{\url{https://share.streamlit.io/johngiorgi/seq2rel/main/demo.py}}}.
\end{abstract}

\section{Introduction}

PubMed, the largest repository of biomedical literature, contains over 30 million publications and is adding more than two papers per minute. Accurate, automated text mining and natural language processing (NLP) methods are needed to maximize discovery and extract structured information from this massive volume of text. An important step in this process is relation extraction (RE), the task of identifying groups of entities within some text that participate in a semantic relationship. In the domain of biomedicine, relations of interest include chemical-induced disease, protein-protein interactions, and gene-disease associations.

Many methods have been proposed for RE, ranging from rule-based to machine learning-based \citep{zhou2014biomedical, DBLP:journals/corr/LiuCJY16}. Most of this work has focused on \textit{intra}-sentence binary RE, where pairs of entities within a sentence are classified as belonging to a particular relation (or none). These methods often ignore commonly occurring complexities like nested or discontinuous entities, coreferent mentions (words or phrases in the text that refer to the same entity), inter-sentence and \(n\)-ary relations (see \autoref{fig:complexities} for examples). The decision not to model these phenomena is a strong assumption. In GENIA \citep{Kim2003GENIAC}, a corpus of PubMed articles labelled with around 100,000 biomedical entities, \(\sim\)17\% of all entities are nested within another entity. Discontinuous entities are particularly common in clinical text, where \(\sim\)10\% of mentions in popular benchmark corpora are discontinuous \citep{wang-etal-2021-discontinuous}. In the CDR corpus \citep{li2016biocreative}, which comprises 1500 PubMed articles annotated for chemical-induced disease relations, \(\sim\)30\% of all relations are inter-sentence. Some relations, like drug-gene-mutation interactions, are difficult to model with binary RE \citep{zhou2014biomedical}.

\begin{figure*}[t]
\centering
\caption{Examples of complexities in entity and relation extraction and the proposed linearization schema to model them. \texttt{CID}: chemical-induced disease. \texttt{GDA}: gene-disease association. \texttt{DGM}: drug-gene-mutation.}
\label{fig:complexities}
\small
\begin{tabular}{p{0.10\textwidth} p{0.58\textwidth} p{0.22\textwidth}}
\toprule
Complexities &
  Example &
  Comment \\ \midrule
Discontinuous mentions &
  Induction by {\color{MyGreen}\textbf{paracetamol}} of {\color{MyOrange}\textbf{bladder}} and {\color{MyOrange} \textbf{liver tumours}}. &
  Discontinuous mention of {\color{MyOrange}\textbf{bladder tumours}}. \\ \cmidrule(lr){2-2}
 &
  \texttt{{\color{MyGreen}\textbf{paracetamol} @DRUG@} {\color{MyOrange}\textbf{bladder tumours} @DISEASE@} @CID@ \newline {\color{MyGreen}\textbf{paracetamol} @DRUG@} {\color{MyOrange}\textbf{liver tumours} @DISEASE@} @CID@} &
   \\ \cmidrule(r){1-3}
Coreferent mentions &
  Proto-oncogene {\color{MyPurple} \textbf{HER2}} (also known as {\color{MyPurple} \textbf{erbB-2}} or {\color{MyPurple} \textbf{neu}}) plays an important role in the carcinogenesis and the prognosis of {\color{MyOrange} \textbf{breast cancer}}. &
  Two coreferent mentions of {\color{MyPurple} \textbf{HER2}}. \\ \cmidrule(lr){2-2}
 &
  \texttt{{\color{MyPurple} \textbf{her2}} ; {\color{MyPurple} \textbf{erbb-2}} ; {\color{MyPurple} \textbf{neu}} {\color{MyPurple} @GENE@} {\color{MyOrange} \textbf{breast cancer} @DISEASE@} @GDA@} &
   \\ \cmidrule(r){1-3}
\(n\)-ary, inter-sentence &
  The deletion mutation on exon-19 of {\color{MyPurple} \textbf{EGFR}} gene was present in 16 patients, while the {\color{MyPurple} \textbf{L858E}} point mutation on exon-21 was noted in 10. All patients were treated with {\color{MyGreen}\textbf{gefitinib}} and showed a partial response. &
  Ternary {\color{MyGreen}\textbf{D}}{\color{MyPurple}\textbf{G}}{\color{MyBlue}\textbf{M}} relationship crosses a sentence boundary. \\ \cmidrule(lr){2-2}
 &
  \texttt{{\color{MyGreen} \textbf{gefitinib} @DRUG@} {\color{MyPurple} \textbf{egfr} @GENE@} {\color{MyBlue} \textbf{l858e} @MUTATION@} @DGM@} &
   \\ \bottomrule
\end{tabular}
\vspace{-3mm}
\end{figure*}

In response to some of these shortcomings, there has been a growing interest in \textit{document}-level RE (DocRE). DocRE aims to model \textit{inter}-sentence relations between coreferent mentions of entities in a document. A popular approach involves graph-based methods, which have the advantage of naturally modelling inter-sentence relations \citep{peng2017cross, song-etal-2018-n, christopoulou-etal-2019-connecting, nan-etal-2020-reasoning, minh-tran-etal-2020-dots}. However, like all pipeline-based approaches, these methods assume that the entities within the text are known. As previous work has demonstrated, and as we show in \textsection \ref{pipeline-vs-end-to-end}, jointly learning to extract entities and relations can improve performance \citep{miwa2014modeling, miwa2016end, gupta2016table, li2016joint, li2017neural, nguyen2019, yu2020joint} and may be more efficient due to shared parameters and training steps. Existing end-to-end methods typically combine task-specific components for entity detection, coreference resolution, and relation extraction that are trained jointly. Most approaches are restricted to intra-sentence RE \citep{bekoulis2018joint, Luan2018MultiTaskIO, nguyen2019end, wadden-etal-2019-entity, giorgi2019end} and have only recently been extended to DocRE \citep{eberts-ulges-2021-end}. However, they still focus on binary relations. Ideally, DocRE methods would be capable of modelling the complexities mentioned above without strictly requiring entities to be known.

A less popular end-to-end approach is to frame RE as a \textit{generative} task with sequence-to-sequence (seq2seq) learning \citep{sutskever2014sequence}. This framing simplifies RE by removing the need for task-specific components and explicit negative training examples, i.e. pairs of entities that \textit{do not} express a relation. If the information to extract is appropriately linearized to a string, seq2seq methods are flexible enough to model all complexities discussed thus far. However, existing work stops short, focusing on intra-sentence binary relations \citep{zeng-etal-2018-extracting, Zhang2020MinimizeEB, nayak2020effective, Zeng2020CopyMTLCM}. In this paper, we extend work on seq2seq methods for RE to the document level, with several important contributions:

\begin{itemize}
    \item We propose a novel linearization schema that can handle complexities overlooked by previous seq2seq approaches, like coreferent mentions and \(n\)-ary relations (\textsection \ref{linearization}).
    \item Using this linearization schema, we demonstrate that a seq2seq approach is able to learn the subtasks of DocRE (entity extraction, coreference resolution and relation extraction) jointly, and report the first end-to-end results on several popular biomedical datasets (\textsection\ref{comparison-to-existing-methods}).
    \item{We devise a simple strategy, referred to as \say{entity hinting} (\textsection\ref{entity-hinting}), to compare our model to existing pipeline-based approaches, in some cases exceeding their performance (\textsection\ref{comparison-to-existing-methods}).}
\end{itemize}

\begin{figure*}[t]
\centering
\includegraphics[width=\linewidth]{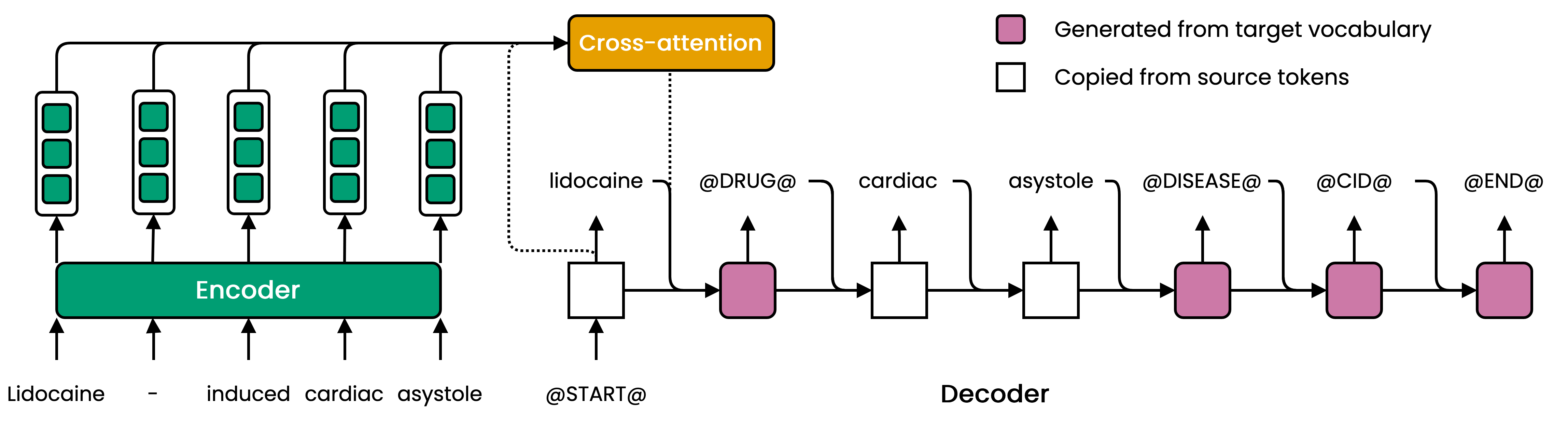}
\caption{A sequence-to-sequence model for document-level relation extraction. Special tokens are generated by the decoder. Entity mentions are copied from the input via a copy mechanism (not shown). Decoding is initiated by a \texttt{@START@} token and terminated when the model generates the \texttt{@END@} token. Attention connections shown only for the second timestep to reduce clutter. \texttt{CID}: chemical-induced disease.}
\label{fig:main}
\end{figure*}

\section{Task definition: document-level relation extraction}

Given a source document of \(S\) tokens, a model must extract all tuples corresponding to a relation, \(R\), expressed between the entities, \(E\) in the document, \((E_1, ..., E_n, R)\) where \(n\) is the number of participating entities, or \textit{arity}, of the relation. Each entity \(E_i\) is represented as the set of its coreferent mentions \(\{e^i_j\}\) in the document, which are often expressed as aliases, abbreviations or acronyms. All entities appearing in a tuple have at least one mention in the document. The mentions that express a given relation are not necessarily contained within the same sentence. Commonly, \(E\) is assumed to be known and provided as input to a model. We will refer to these methods as \say{pipeline-based}. In this paper, we are primarily concerned with the situation where \(E\) is \textit{not} given and must be predicted by a model, which we will refer to as \say{end-to-end}.

\section{Our approach: seq2rel}

\subsection{Linearization} \label{linearization}

To use seq2seq learning for RE, the information to be extracted must be linearized to a string. This linearization should be expressive enough to model the complexities of entity and relation extraction without being overly verbose. We propose the following schema, illustrated with an example: \\

\noindent
{\small
\(X\): Variants in the {\color{MyPurple} \textbf{estrogen receptor alpha}} ({\color{MyPurple} \textbf{ESR1}}) gene and its mRNA contribute to risk for {\color{MyOrange} \textbf{schizophrenia}}.
}\\
\noindent
{\small
\(Y\): \texttt{{\color{MyPurple} \textbf{estrogen receptor alpha}} ; {\color{MyPurple} \textbf{ESR1} @GENE@} {\color{MyOrange} \textbf{schizophrenia} @DISEASE@} @GDA@}
}\\

\noindent
The input text \(X\), expresses a gene-disease association (GDA) between {\color{MyPurple} \textbf{ESR1}} and {\color{MyOrange} \textbf{schizophrenia}}. In the corresponding target string \(Y\), each relation begins with its constituent entities. A semicolon separates coreferent mentions (\texttt{;}), and entities are terminated with a special token denoting their type (e.g. \texttt{@GENE@}). Similarly, relations are terminated with a special token denoting their type (e.g. \texttt{@GDA@}). Two or more entities can be included before the special relation token to support \(n\)-ary extraction. Entities can be ordered if they serve specific roles as head or tail of a relation. For each document, multiple relations can be included in the target string. Entities may be nested or discontinuous in the input text. In \autoref{fig:complexities}, we provide examples of how this schema can be used to model various complexities, like coreferent entity mentions and \(n\)-ary relations.

\subsection{Model} \label{model}

The model follows a canonical seq2seq setup. An encoder maps each token in the input to a contextual embedding. An autoregressive decoder generates an output, token-by-token, attending to the outputs of the encoder at each timestep (\autoref{fig:main}). Decoding proceeds until a special \say{end-of-sequence} token (\texttt{@END@}) is generated, or a maximum number of tokens have been generated. Formally, \(X\) is the \textit{source} sequence of length \(S\), which is some text we would like to extract relations from. \(Y\) is the corresponding \textit{target} sequence of length \(T\), a linearization of the relations contained in the source. We model the conditional probability

\begin{align}
    p(Y | X) = \prod_{t=1}^Tp(y_t | X, y_{<t})    
\end{align}

\noindent During training, we optimize over the model parameters \(\theta\) the sequence cross-entropy loss

\begin{align}
    \ell(\theta) = - \sum_{t=1}^T \log p(y_t | X, y_{<t} ; \theta) \label{cross-entropy-loss}
\end{align}

\noindent maximizing the log-likelihood of the training data.\footnote{See \textsection \ref{implementation} for details about the encoder and decoder.} 

The main problems with this setup for RE are: 1) The model might \say{hallucinate} by generating entity mentions that do not appear in the source text. 2) It may generate a target string that does not follow the linearization schema and therefore cannot be parsed. 3) The loss function is permutation-sensitive, enforcing an unnecessary decoding order. To address 1) we use two modifications: a restricted target vocabulary (\textsection\ref{restricted-target-vocabulary}) and a copy mechanism (\textsection\ref{copy-mechanism}). To address 2) we experiment with several constraints applied during decoding (\textsection\ref{constrained-decoding}). Finally, to address 3) we sort relations according to their order of appearance in the source text (\textsection\ref{sorting-relations}).

\subsubsection{Restricted target vocabulary} \label{restricted-target-vocabulary}

To prevent the model from \say{hallucinating} (generating entity mentions that do not appear in the source text), the target vocabulary is restricted to the set of special tokens needed to model entities and relations (e.g. \texttt{;} and \texttt{@DRUG@}). All other tokens must be copied from the input using a copy mechanism (see \textsection\ref{copy-mechanism}). The embeddings of these special tokens are initialized randomly and learned jointly with the rest of the model's parameters.

\subsubsection{Copy mechanism} \label{copy-mechanism}

To enable copying of input tokens during decoding, we use a copying mechanism \citep{gu2016incorporating}. The mechanism works by effectively extending the target vocabulary with the tokens in the source sequence \(X\), allowing the model to \say{copy} these tokens into the output sequence, \(Y\). Our use of the copy mechanism is similar to previous seq2seq-based approaches for RE \citep{zeng-etal-2018-extracting, Zeng2020CopyMTLCM}.

\subsubsection{Constrained decoding} \label{constrained-decoding}

We experimented with several constraints applied to the decoder during test time to reduce the likelihood of generating syntactically invalid target strings (strings that do not follow the linearization schema). These constraints are applied by setting the predicted probabilities of invalid tokens to a tiny value at each timestep. The full set of constraints is depicted in \autoref{appendix:constrained-decoding}. In practice, we found that a trained model rarely generates invalid target strings, so these constraints have little effect on final performance (see \textsection\ref{ablation}). We elected not to apply them in the rest of our experiments.

\subsubsection{Sorting relations} \label{sorting-relations}

The relations to extract from a given document are inherently unordered. However, the sequence cross-entropy loss (\autoref{cross-entropy-loss}) is permutation-sensitive with respect to the predicted tokens. During training, this enforces an unnecessary decoding order and may make the model prone to overfit frequent token combinations in the training set \citep{Vinyals2016OrderMS, yang2019deep}. To partially mitigate this, we sort relations within the target strings according to their order of appearance in the source text, providing the model with a consistent decoding order. The position of a relation is determined by the first occurring mention of its head entity. The position of a mention is determined by the sum of its start and end character offsets. In the case of ties, we then sort by the first mention of its tail entity (and so on for \(n\)-ary relations).

\subsection{Entity hinting} \label{entity-hinting}

Although the proposed model can jointly extract entities and relations from unannotated text, most existing DocRE methods provide the entities as input. Therefore, to more fairly compare to existing methods, we also provide entities as input, using a simple strategy that we will refer to as \say{entity hinting}. This involves prepending entities to the source text as they appear in the target string. Taking the example from \textsection\ref{linearization}, entity hints would be added as follows:\\

\noindent
{\small
\(X\): \texttt{{\color{MyPurple} \textbf{estrogen receptor alpha}} ; {\color{MyPurple} \textbf{ESR1} @GENE@} {\color{MyOrange} \textbf{schizophrenia} @DISEASE@} @SEP@} Variants in the {\color{MyPurple} \textbf{estrogen receptor alpha}} ({\color{MyPurple} \textbf{ESR1}}) gene and its mRNA contribute to risk for {\color{MyOrange} \textbf{schizophrenia}}.
}
\\

\noindent where the special \texttt{@SEP@} token demarcates the end of the entity hint.\footnote{Some pretrained models have their own separator token which can be used in place of \texttt{@SEP@}, e.g. BERT uses \texttt{[SEP]}.} We experimented with the common approach of inserting marker tokens before and after each entity mention \citep{Zhou2021AnIB} but found this to perform worse. Our approach adds fewer extra tokens to the source text and provides a location for the copy mechanism to focus, i.e. tokens left of \texttt{@SEP@}. In our experiments, we use entity hinting when comparing to methods that provide ground truth entity annotations as input (\textsection \ref{existing-pipeline-based-methods}). In \textsection \ref{pipeline-vs-end-to-end}, we use entity hinting to compare pipeline-based and end-to-end approaches.

\section{Experimental setup}

\subsection{Datasets}

We evaluate our approach on several biomedical, DocRE datasets. We also include one non-biomedical dataset, DocRED. In \autoref{appendix:dataset-details}, we list relevant details about their annotations.

\paragraph{CDR \citep{li2016biocreative}}

The BioCreative V CDR task corpus is manually annotated for chemicals, diseases and chemical-induced disease (CID) relations. It contains the titles and abstracts of 1500 PubMed articles and is split into equally sized train, validation and test sets. Given the relatively small size of the training set, we follow \citet{christopoulou-etal-2019-connecting} and others by first tuning the model on the validation set and then training on the combination of the train and validation sets before evaluating on the test set. Similar to prior work, we filter negative relations with disease entities that are hypernyms of a corresponding true relations disease entity within the same abstract (see \autoref{appendix:hypernym-filtering}).

\paragraph{GDA \citep{renet2019}}

The gene-disease association corpus contains 30,192 titles and abstracts from PubMed articles that have been automatically labelled for genes, diseases and gene-disease associations via distant supervision. The test set is comprised of 1000 of these examples. Following \citet{christopoulou-etal-2019-connecting} and others, we hold out a random 20\% of the remaining abstracts as a validation set and use the rest for training.

\paragraph{DGM \citep{jia-etal-2019-document}}

The drug-gene-mutation corpus contains 4606 PubMed articles that have been automatically labelled for drugs, genes, mutations and ternary drug-gene-mutation relationships via distant supervision. The dataset is available in three variants: sentence, paragraph, and document-length text. We train and evaluate our model on the paragraph-length inputs. Since the test set does not contain relation annotations on the paragraph level, we report results on the validation set. We hold out a random 20\% of training examples to form a new validation set for tuning.

\paragraph{DocRED \citep{Yao2019DocREDAL}}

DocRED includes over 5000 human-annotated documents from Wikipedia. There are six entity and 96 relation types, with \(\sim\)40\% of relations crossing the sentence boundary. We use the same split as previous end-to-end methods \citep{eberts-ulges-2021-end}, which has 3,008 documents in the training set, 300 in the validation set and 700 in the test set\footnote{\url{https://github.com/lavis-nlp/jerex}}.

\subsection{Evaluation}

We evaluate our model using the micro F1-score by extracting relation tuples from the decoder's output (see \autoref{appendix:parsing-model-output}). Similar to prior work, we use a \say{strict} criteria. A predicted relation is considered correct if the relation type and its entities match a ground truth relation. An entity is considered correct if the entity type and its mentions match a ground truth entity. However, since the aim of DocRE is to extract relations at the \textit{entity}-level (as opposed to the \textit{mention}-level), we also report performance using a relaxed criterion (denoted \say{relaxed}), where predicted entities are considered correct if more than 50\% of their mentions match a ground truth entity (see \autoref{appendix:relaxed-entity-matching}).

Existing methods that evaluate on CDR, GDA and DGM use the ground truth entity annotations as input. This makes it difficult to directly compare with our end-to-end approach, which takes only the raw text as input. To make the comparison fairer, we use entity hinting (\textsection \ref{entity-hinting}) so that our model has access to the ground truth entity annotations. We also report the performance of our method in the end-to-end setting on these corpora to facilitate future comparison. To compare to existing end-to-end approaches, we use DocRED.

\subsection{Implementation, training and hyperparameters}

\paragraph{Implementation} \label{implementation}
We implemented our model in PyTorch \citep{pytorch} using AllenNLP \citep{AllenNLP}. As encoder, we use a pretrained transformer, implemented in the Transformers library \citep{hf-transformers}, which is fine-tuned during training. When training and evaluating on biomedical corpora, we use PubMedBERT \citep{gu2020domain}, and BERT\textsubscript{BASE} \citep{devlin-etal-2019-bert} otherwise. In both cases, we use the default hyperparameters of the pretrained model. As decoder, we use a single-layer LSTM \citep{Hochreiter1997LongSM} with randomly initialized weights. We use multi-head attention \citep{vaswani2017attention} as the cross-attention mechanism between encoder and decoder. Select hyperparameters were tuned on the validation sets, see \autoref{appendix:hyperparameters} for details.

\paragraph{Training} \label{training}

All parameters are trained jointly using the AdamW optimizer \citep{adamw}. Before training, we re-initialize the top \(L\) layers of the pretrained transformer encoder, which has been shown to improve performance and stability during fine-tuning \citep{Zhang2021RevisitingFB}. During training, the learning rate is linearly increased for the first 10\% of training steps and linearly decayed to zero afterward. Gradients are scaled to a vector norm of \(1.0\) before backpropagating. During each forward propagation, the hidden state of the LSTM decoder is initialized with the mean of token embeddings output by the encoder. The decoder is regularized by applying dropout \citep{Srivastava2014DropoutAS} with probability \(0.1\) to its inputs, and DropConnect \citep{Wan2013RegularizationON} with probability \(0.5\) to the hidden-to-hidden weights. As is common, we use teacher forcing, feeding previous ground truth inputs to the decoder when predicting the next token in the sequence. During test time, we generate the output using beam search \citep{graves2012sequence}. Beams are ranked by mean token log probability after applying a length penalty.\footnote{\url{https://docs.allennlp.org/main/api/nn/beam_search/\#lengthnormalizedsequencelogprobabilityscorer}} Models were trained and evaluated on a single NVIDIA Tesla V100.\footnote{\url{https://www.nvidia.com/en-us/data-center/v100/}}

\section{Results}

\subsection{Comparison to existing methods} \label{comparison-to-existing-methods}

In the following sections, we compare our model to existing DocRE methods on several benchmark corpora. We compare to existing pipeline-based methods (\textsection \ref{existing-pipeline-based-methods}), including \(n\)-ary methods (\textsection \ref{existing-n-ary-methods}), and end-to-end methods (\textsection \ref{existing-end-to-end-methods}). Details about these methods are provided in \autoref{appendix:baselines}.

\begin{table}[t]
\centering
\caption{Comparison to existing pipeline-based methods. Performance reported as micro-precision, recall and F1-scores (\%) on the CDR and GDA test sets. Results below the horizontal line are not comparable to existing methods. Bold: best scores.}
\label{tab:02}
\resizebox{\columnwidth}{!}{%
\begin{tabular}{@{}lcccccc@{}}
\toprule
                                           & \multicolumn{3}{c}{CDR}     & \multicolumn{3}{c}{GDA}     \\ \midrule
Method                                     & P    & R    & F1            & P    & R    & F1            \\ \midrule
\citet{christopoulou-etal-2019-connecting} & 62.1 & 65.2 & 63.6          & --   & --   & 81.5          \\
\citet{nan-etal-2020-reasoning}            & --   & --   & 64.8          & --   & --   & 82.2          \\
\citet{minh-tran-etal-2020-dots}           & --   & --   & 66.1          & --   & --   & 82.8          \\
\citet{lai2020bert}                        & 64.9 & 67.1 & 66.0          & --   & --   & --            \\
\citet{Xu2021EntitySW}                     & --   & --   & 68.7          & --   & --   & 83.7          \\
\citet{zhou2021document}                   & --   & --   & \textbf{69.4} & --   & --   & 83.9          \\
seq2rel (entity hinting)                   & 68.2 & 66.2 & 67.2          & 84.4 & 85.3 & \textbf{84.9} \\ \midrule
seq2rel (entity hinting, relaxed)          & 68.2 & 66.2 & 67.2          & 84.5 & 85.4 & 85.0          \\
seq2rel (end-to-end)                       & 43.5 & 37.5 & 40.2          & 55.0 & 55.4 & 55.2          \\
seq2rel (end-to-end, relaxed)              & 56.6 & 48.8 & 52.4          & 70.3 & 70.8 & 70.5          \\ \bottomrule
\end{tabular}%
}
\vspace{-1mm}
\end{table}

\begin{figure}[t]
\centering
\includegraphics[width=\linewidth]{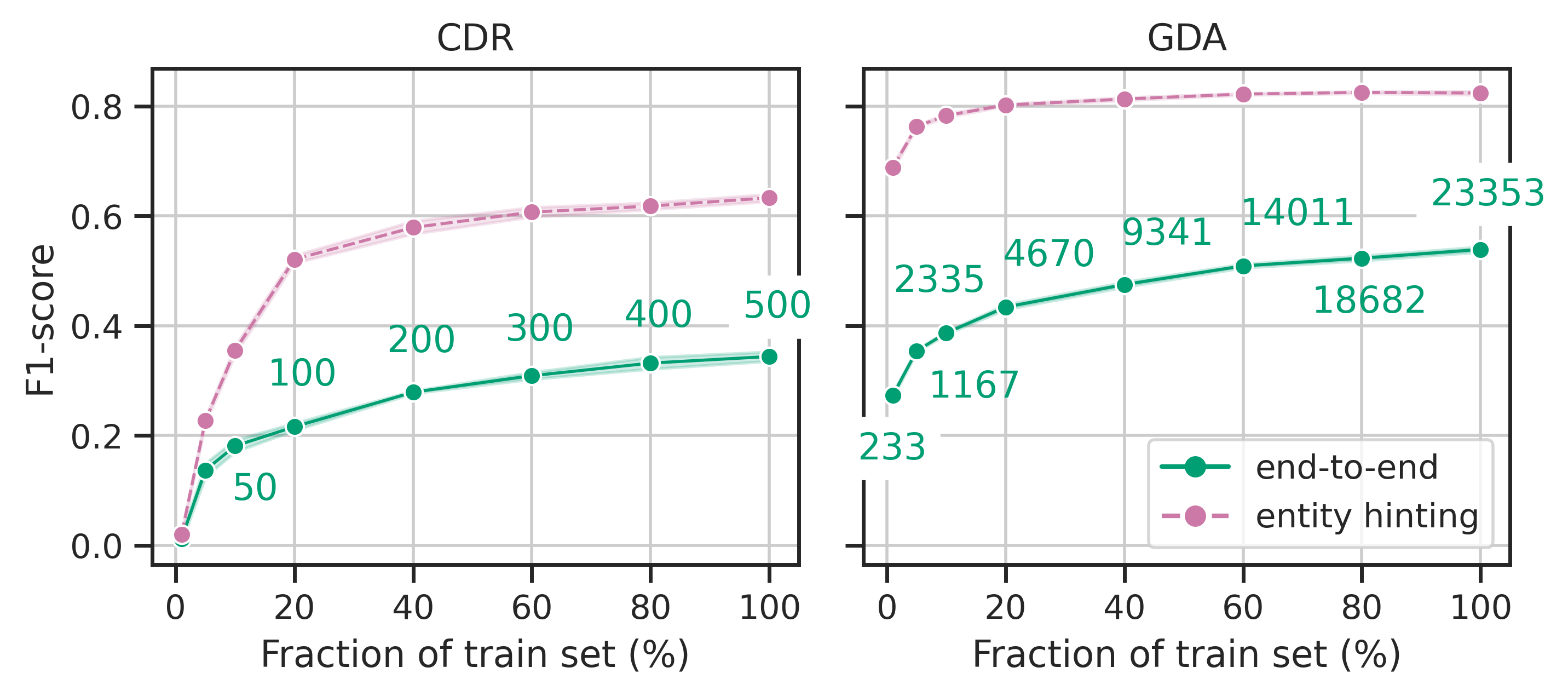}
\caption{Effect of training set size on performance. Performance reported as the median micro F1-score obtained over five runs with different random seeds on the CDR and GDA validation sets, with and without entity hinting. Error bands correspond to the standard deviation over the five runs. The absolute number of training examples are displayed for each corpus. Some labels are excluded to reduce clutter.}
\label{fig:dataset-size}
\end{figure}

\subsubsection{Existing pipeline-based methods}
\label{existing-pipeline-based-methods}

In \autoref{tab:02}, we use entity hinting to compare our method to existing pipeline-based methods on CDR and GDA. We also report end-to-end performance, which is not comparable to existing pipeline-based methods but will facilitate future comparisons.

The large performance improvement when using entity hinting (+27-29\%) confirms that the model exploits the entity annotations. The fact that relaxed entity matching makes a large difference in the end-to-end setting (+12-15\%) suggests that a significant portion of the model's mistakes occur during coreference resolution. Although our method is designed for end-to-end RE, we find that it outperforms existing pipeline-based methods when using entity hinting on GDA. Our method is competitive with existing methods when using entity hinting on the CDR corpus but ultimately underperforms state-of-the-art results. Given that GDA is 46X larger, we speculated that our method might be underperforming in the low-data regime. To determine if this is a contributing factor, we artificially reduce the size of the CDR and GDA training sets and plot the performance as a curve (\autoref{fig:dataset-size}). In all cases besides GDA with entity hinting, performance increases monotonically with dataset size. There is no obvious plateau on CDR even when using all 500 training examples. Together, these results suggest that our seq2seq based approach can outperform existing pipeline-based methods when there are sufficient training examples but underperforms relative to existing methods in the low-data regime.

\begin{table}[t]
\centering
\caption{Comparison to existing \(n\)-ary methods. Performance reported as micro-precision, recall and F1-scores (\%) on the DGM validation set. Results below the horizontal line are not comparable to existing methods. Bold: best scores. \textsuperscript{\textdagger} \citealt{jia-etal-2019-document} do not report results on the validation set, so we re-run their paragraph-level model.}
\label{tab:03}
\small
\begin{tabular}{@{}lccc@{}}
\toprule
Method                                                       & P             & R             & F1            \\ \midrule
\citet{jia-etal-2019-document} \textsuperscript{\textdagger} & 62.9          & 76.2          & 68.9          \\
seq2rel (entity hinting)                                     & \textbf{84.0} & \textbf{84.8} & \textbf{84.4} \\ \midrule
seq2rel (entity hinting, relaxed)                            & 84.1          & 84.9          & 84.5          \\
seq2rel (end-to-end)                                         & 68.9          & 65.9          & 67.4          \\
seq2rel (end-to-end, relaxed)                                & 78.3          & 74.9          & 76.6          \\ \bottomrule
\end{tabular}
\vspace{-3mm}
\end{table}

\subsubsection[n-ary relation extraction]{\(n\)-ary relation extraction}
\label{existing-n-ary-methods}

In \autoref{tab:03} we compare to existing \(n\)-ary methods on the DGM corpus. With entity hinting, our method significantly outperforms the existing method. The difference in encoders partially explains this large performance gap. Where \citet{jia-etal-2019-document} use a BiLSTM that is trained from scratch, we use PubMedBERT, a much larger model that has been pretrained on abstracts and full-text articles from PubMedCentral.\footnote{\url{https://www.ncbi.nlm.nih.gov/pmc/}} However, this does not completely account for the improvement in performance, as recent work that has replaced the BiLSTM encoder of \citep{jia-etal-2019-document} with PubMedBERT found that it improves performance by approximately 2-4\% on the task of drug-gene-mutation prediction \citep{zhang-etal-2021-modular}.\footnote{The authors have not released code at the time of writing, so we were unable to evaluate this model on the DGM validation set in order to compare with our method directly.} Our results on the DGM corpus suggest that our linearization schema effectively models \(n\)-ary relations without requiring changes to the model architecture or training procedure.

\subsubsection{End-to-end methods}
\label{existing-end-to-end-methods}

In \autoref{tab:04} we compare to an existing end-to-end approach on DocRED, JEREX \citep{eberts-ulges-2021-end}. To make the comparison fair, we use the same pretrained encoder (BERT\textsubscript{BASE}). We find that
although our model is arguably simpler (JEREX contains four task-specific sub-components, each with its own loss)
it only slightly underperforms JEREX, mainly due to recall. We speculate that one reason for this is a large number of relations per document, which leads to longer target strings and, therefore, more decoding steps. The median length of the target strings in DocRED, using our linearization, is 110, whereas the next largest is 19 in GDA. Improving the decoder's ability to process long sequences, e.g. switching the LSTM for a transformer or modifying the linearization schema to produce shorter target strings, may improve recall and close the gap with existing methods.

\begin{table}[t]
\centering
\caption{Comparison to existing end-to-end methods. Performance reported as micro-precision, recall and F1-scores (\%) on the DocRED test set. Results below the horizontal line are not comparable to existing methods. Bold: best scores.}
\label{tab:04}
\small
\begin{tabular}{@{}lccc@{}}
\toprule
Method                              & P             & R             & F1            \\ \midrule
JEREX \citep{eberts-ulges-2021-end} & 42.8          & \textbf{38.2} & \textbf{40.4} \\
seq2rel (end-to-end)                & \textbf{44.0} & 33.8          & 38.2          \\ \midrule
seq2rel (end-to-end, relaxed)       & 53.7          & 41.3          & 46.7          \\ \bottomrule
\end{tabular}
\end{table}

\begin{table}[t]
\centering
\caption{Comparison of pipeline-based and end-to-end approaches. Gold hints use gold-standard entity annotations to insert entity hints in the source text. Silver hints use the entity annotations provided by PubTator. Pipeline is identical to silver entity hints, except that we filter out entity mentions predicted by our model that PubTator does not predict. The end-to-end model only has access to the unannotated source text as input. Performance reported as micro-precision, recall and F1-scores (\%) on the CDR test set, with strict and relaxed entity matching criteria. Bold: best scores.}
\label{tab:05}
\small
\begin{tabular}{@{}lcccccc@{}}
\toprule
             & \multicolumn{3}{c}{Strict}                    & \multicolumn{3}{c}{Relaxed}                   \\ \midrule
             & P             & R             & F1            & P             & R             & F1            \\ \cmidrule(l){2-7} 
Gold hints   & 68.2          & 66.2          & 67.2          & 68.2          & 66.2          & 67.2          \\ \midrule
Silver hints & 42.4          & 37.3          & 39.7          & 53.0          & 46.7          & 49.7          \\
Pipeline     & \textbf{45.0} & 16.9          & 24.6          & \textbf{62.5} & 23.5          & 34.1          \\
End-to-end   & 43.5          & \textbf{37.5} & \textbf{40.2} & 56.6          & \textbf{48.8} & \textbf{52.4} \\ \bottomrule
\end{tabular}
\vspace{-2mm}
\end{table}

\subsection{Pipeline vs. End-to-end} \label{pipeline-vs-end-to-end}

In \textsection \ref{existing-pipeline-based-methods} and \textsection \ref{existing-n-ary-methods}, we provide gold-standard entity annotations from each corpus as input to our model via entity hinting (referred to as \say{gold} hints from here on, see \textsection \ref{entity-hinting}). This allowed us to compare to existing methods that also provide these annotations as input. However, gold-standard entity annotations are (almost) never available in real-world settings, such as large-scale extraction on PubMed. In this setting, there are two strategies: pipeline-based, where independent systems perform entity and relation extraction, and end-to-end, where a single model performs both tasks. To compare these approaches under our model, we perform evaluations where a named entity recognition (NER) system is used to determine entity hints (referred to as \say{silver} hints from here on) and when no entity hints are provided (end-to-end).\footnote{Specifically, we use PubTator \citep{wei2013pubtator}. PubTator provides up-to-date entity annotations for PubMed using state-of-the-art machine learning systems.}  However, this alone does not create a true pipeline, as our model can recover from both false negatives and false positives in the NER step. To mimic error propagation in the pipeline setting, we filter any entity mention predicted by our model that was \textit{not} predicted by the NER system. In \autoref{tab:05}, we present the results of all four settings (gold and silver entity hints, pipeline and end-to-end) on CDR.

We find that using gold entity hints significantly outperforms all other settings. This is expected, as the gold-standard entity annotations are high-quality labels produced by domain experts. Using silver hints significantly drops performance, likely due to a combination of false positive and false negatives from the NER step. In the pipeline setting, where there is no recovery from false negatives, performance falls by another 15\%. The end-to-end setting significantly outperforms the pipeline setting (due to a large boost in recall) and performs comparably to using silver hints. Together, our results suggest that performance reported using gold-standard entity annotations may be overly optimistic and corroborates previous work demonstrating the benefits of jointly learning entity and relation extraction \citep{miwa2014modeling, miwa2016end, gupta2016table, li2016joint, li2017neural, nguyen2019, yu2020joint}.

\begin{table}[t]
\centering
\caption{Ablation study results. Performance reported as the micro-precision, recall and F1-scores (\%) on the CDR and DocRED validation sets. \(\Delta\): difference to the complete models F1-score. Bold: best scores.}
\label{tab:06}
\resizebox{\columnwidth}{!}{%
\begin{tabular}{@{}lllllllll@{}}
\toprule
                       & \multicolumn{4}{c}{CDR}                      & \multicolumn{4}{c}{DocRED}          \\ \midrule
 &
  \multicolumn{1}{c}{P} &
  \multicolumn{1}{c}{R} &
  \multicolumn{1}{c}{F1} &
  \multicolumn{1}{c}{\(\Delta\)} &
  \multicolumn{1}{c}{P} &
  \multicolumn{1}{c}{R} &
  \multicolumn{1}{c}{F1} &
  \multicolumn{1}{c}{\(\Delta\)} \\ \cmidrule(l){2-9} 
seq2rel (end-to-end) &
  \textbf{41.0} &
  35.1 &
  37.8 &
  \multicolumn{1}{c}{--} &
  46.9 &
  \textbf{36.1} &
  \textbf{40.8} &
  \multicolumn{1}{c}{--} \\ \midrule
- pretraining          & 9.4  & 6.9           & 8.0           & -29.8 & 18.5          & 7.7  & 10.8 & -30.0 \\
- fine-tuning          & 24.3 & 20.5          & 22.2          & -15.6 & 42.4          & 15.5 & 22.7 & -18.1 \\
- vocab restriction    & 39.6 & 32.2          & 35.5          & -2.3  & 45.2          & 35.5 & 39.7 & -1.1  \\
- sorting relations    & 36.1 & 29.2          & 32.3          & -5.6  & \textbf{52.9} & 17.4 & 26.2 & -14.7 \\
+ constrained decoding & 40.8 & \textbf{35.6} & \textbf{38.0} & +0.2  & 46.8          & 35.9 & 40.6 & -0.2  \\ \bottomrule
\end{tabular}%
}
\vspace{-4mm}
\end{table}

\subsection{Ablation} \label{ablation}

In \autoref{tab:06}, we present the results of an ablation study. We perform the analysis twice, once on the biomedical corpus CDR and once on the general domain corpus DocRED. Unsurprisingly, we find that fine-tuning a pretrained encoder greatly impacts performance. Training the same encoder from scratch (- pretraining) reduces performance by \(\sim\)30\%. Using the pretrained weights without fine-tuning (- fine-tuning) drops performance by 15.6-18.1\%. Restricting the target vocabulary (- vocab restriction, see \textsection \ref{restricted-target-vocabulary}) has a small positive impact, boosting performance by 1.1\%-2.3\%. Deliberately ordering the relations within each target string (- sorting relations, see \textsection\ref{sorting-relations}) has a large positive impact, boosting performance by 5.6\%-14.7\%. This effect is larger on DocRED, likely because it has more relations per document on average than CDR, so ordering becomes more important. Finally, adding constraints to the decoding process (+ constrained decoding) has little impact on performance, suggesting that a trained model rarely generates invalid target strings (see \textsection \ref{constrained-decoding}).

\section{Discussion}

\subsection{Related work}

Seq2seq learning for RE has been explored in prior work. CopyRE \citep{zeng-etal-2018-extracting} uses an encoder-decoder architecture with a copy mechanism, similar to our approach, but is restricted to intra-sentence relations. Additionally, because CopyRE's decoding proceeds for exactly three timesteps per relation, the model is limited to generating binary relations between single token entities. The ability to decode multi-token entities was addressed in follow-up work, CopyMTL \citep{Zeng2020CopyMTLCM}. A similar approach was published concurrently but was again limited to intra-sentence binary relations \citep{nayak2020effective}. Most recently, GenerativeRE \citep{cao-ananiadou-2021-generativere-incorporating} proposed a novel copy mechanism to improve performance on multi-token entities. None of these approaches deal with the complexities of DocRE, where many relations cross the sentence boundary, and coreference resolution is critical.\footnote{Concurrent to our work, REBEL \citep{huguet-cabot-navigli-2021-rebel-relation} also extends seq2seq methods to document-level RE, achieving strong performance on DocRED. However, the method was not evaluated on \(n\)-ary relations.}

More generally, our paper is related to a recently proposed \say{text-to-text} framework \citep{raffel2020exploring}. In this framework, a task is formulated so that the inputs and outputs are both text strings, enabling the use of the same model, loss function and even hyperparameters across many seq2seq, classification and regression tasks. This framework has recently been applied to biomedical literature to perform named entity recognition, relation extraction (binary, intra-sentence), natural language inference, and question answering \citep{phan2021scifive}. Our work can be seen as an attempt to formulate the task of DocRE within this framework.

\subsection{Limitations and future work}

\paragraph{Permutation-sensitive loss} Our approach adopts the sequence cross-entropy loss (\autoref{cross-entropy-loss}), which is sensitive to the order of predicted tokens, enforcing an unnecessary decoding order on the inherently unordered relations. To partially mitigate this problem, we order relations within the target string according to order of appearance in the source text, providing the model with a consistent decoding order that can be learned (see \textsection \ref{sorting-relations}, \textsection \ref{ablation}). Previous work has addressed this issue with various strategies, including reinforcement learning \citep{zeng-etal-2019-learning}, unordered-multi-tree decoders \citep{Zhang2020MinimizeEB}, and non-autoregressive decoders \citep{sui2020joint}. However, these works are limited to binary intra-sentence relation extraction, and their suitability for DocRE has not been explored. A promising future direction would be to modify our approach such that the arbitrary order of relations is not enforced during training.

\paragraph{Input length restriction} Due to the pretrained encoder's input size limit (512 tokens), our experiments are conducted on paragraph-length text. Our model could be extended to full documents by swapping its encoder with any of the recently proposed \say{efficient transformers} \citep{tay2021long}. Future work could evaluate such a model's ability to extract relations from full scientific papers.

\paragraph{Pretraining the decoder}

In our model, the encoder is pretrained, while the decoder is trained from scratch. Several recent works, such as T5 \citep{raffel2020exploring} and BART \citep{lewis2019bart}, have proposed pretraining strategies for entire encoder-decoder architectures, which can be fine-tuned on downstream tasks. An interesting future direction would be to fine-tune such a model on DocRE using our linearization schema.

\section{Conclusion}

In this paper, we extend generative, seq2seq methods for relation extraction to the document level. We propose a novel linearization schema that can handle complexities overlooked by previous seq2seq approaches, like coreferent mentions and \(n\)-ary relations. We compare our approach to existing pipeline-based and end-to-end methods on several benchmark corpora, in some cases exceeding their performance. In future work, we hope to extend our method to full scientific papers and develop strategies to improve performance in the low-data regime and in cases where there are many relations per document.

\section*{Acknowledgements}

This research was enabled in part by support provided by Compute Ontario (\href{https://www.computeontario.ca/}{www.computeontario.ca}), Compute Canada (\href{https://www.computecanada.ca/}{www.computecanada.ca}) and the CIFAR AI Chairs Program and partially funded by the US National Institutes of Health (NIH) [U41 HG006623, U41 HG003751).

\bibliography{anthology,custom}

\appendix

\begin{figure*}[t]
\centering
\includegraphics[width=\linewidth]{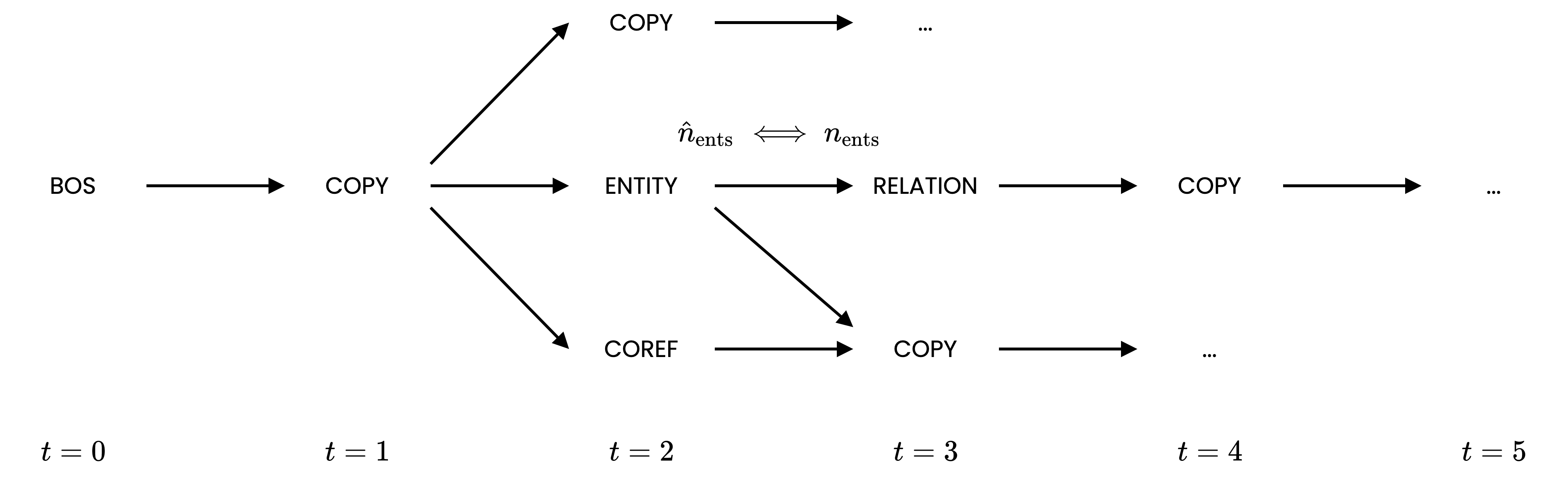}
\caption{A diagram depicting syntactically valid predictions during decoding at each timestep \(t\). The log probabilities of all other possible predictions are set to a tiny value to prevent the model from producing a syntactically invalid target string. \texttt{BOS} is the special beginning-of-sequence token, \texttt{COPY} denotes any token copied from the source text, and \texttt{COREF} is the special token used to separate coreferent mentions (i.e. \texttt{;}). \texttt{ENTITY} is any special entity token (e.g. \texttt{@GENE@}) and \texttt{RELATION} any special relation token (e.g. \texttt{@GDA@} for gene-disease association). \(\hat n_{\text{ents}}\) denotes the number of entities predicted by the current timestep and \(n_{\text{ents}}\) the expected arity of the relation. The special end-of-sequence token (not shown) is always considered valid and its log probability is never modified.}
\label{fig:a1}
\end{figure*}

\section{Constrained decoding}
\label{appendix:constrained-decoding}

In \autoref{fig:a1}, we illustrate the rules used to constrain decoding. At each timestep \(t\), given the prediction of the previous timestep \(t - 1\), the predicted class probabilities of tokens that would generate a syntactically invalid target string are set to a tiny value. In practice, we found that a model rarely generates invalid target strings, so these constraints have little effect on final performance (see \textsection\ref{constrained-decoding} and \textsection\ref{ablation}).

\section{Details about dataset annotations}
\label{appendix:dataset-details}

In \autoref{tab:a1}, we list which complexities (e.g. nested \& discontinuous mentions, \(n\)-ary relations) are contained within each dataset used in our evaluations. We also report the fraction of relations in the test set that are inter-sentence. We consider a relation intra-sentence if \textit{any} sentence in the document contains \textit{at least one} mention of each entity in the relation, and inter-sentence otherwise. This produces an estimate that matches previously reported numbers for CDR (\(\sim\)30\%). In \citet{Yao2019DocREDAL}, the fraction of inter-sentence relations in DocRED is reported as \(\sim\)40.7\%. We can reproduce this value if we consider relations intra-sentence when \textit{all} mentions of an entity exist within a single sentence and inter-sentence otherwise.

\begin{table*}[t]
\centering
\caption{Evaluation datasets used in this paper with details about their annotations. Inter-sentence relations (\%) are the fraction of relations in the test set that cross sentence boundaries. We consider a relation intra-sentence if any sentence in the document contains at least one mention of each entity in the relation, and inter-sentence otherwise. *This differs from the estimate in \citet{Yao2019DocREDAL}, see \autoref{appendix:dataset-details}.}
\label{tab:a1}
\resizebox{\textwidth}{!}{%
\begin{tabular}{@{}lccccc@{}}
\toprule
Corpus                             & Nested Mentions?       & Discontinuous Mentions? & Coreferent mentions?   & \(n\)-ary relations?   & Inter-sentence relations (\%) \\ \midrule
CDR \citep{li2016biocreative}      & \color{MyGreen} \cmark & \color{MyGreen} \cmark  & \color{MyGreen} \cmark & \color{red} \xmark     & 29.8                          \\
GDA \citep{renet2019}          & \color{MyGreen} \cmark & \color{red} \xmark & \color{MyGreen} \cmark & \color{red} \xmark & 15.6 \\
DGM \citep{jia-etal-2019-document} & \color{red} \xmark     & \color{red} \xmark      & \color{MyGreen} \cmark & \color{MyGreen} \cmark & 63.5                          \\
DocRED \citep{Yao2019DocREDAL} & \color{red} \xmark     & \color{red} \xmark & \color{MyGreen} \cmark & \color{red} \xmark & 12.5* \\ \bottomrule
\end{tabular}%
}
\end{table*}

\section{Hypernym filtering}
\label{appendix:hypernym-filtering}

The CDR dataset is annotated for chemical-induced disease (CID) relationships between the most specific chemical and disease mentions in an abstract. Take the following example from the corpus:\\

\noindent
{\small
{\color{MyGreen}\textbf{Carbamazepine}}-induced {\color{MyOrange}\textbf{cardiac dysfunction}} [...] A patient with sinus {\color{MyOrange}\textbf{bradycardia}} and {\color{MyOrange}\textbf{atrioventricular block}}, induced by {\color{MyGreen}\textbf{carbamazepine}}, prompted an extensive literature review of all previously reported cases.
}\\

\noindent In this example (PMID: 1728915), only (\textit{carbamazepine}, \textit{bradycardia}) and (\textit{carbamazepine}, \textit{atrioventricular block}) are labelled as true relations. The relation (\textit{carbamazepine}, \textit{cardiac dysfunction}), although true, is not labelled as \textit{cardiac dysfunction} is a hypernym of both \textit{bradycardia} and \textit{atrioventricular block}. This can harm evaluation performance, as the prediction (\textit{carbamazepine}, \textit{cardiac dysfunction}) will be considered a false positive. Therefore, we follow previous work \citep{Gu2016ChemicalinducedDR, Gu2017ChemicalinducedDR, Verga2018SimultaneouslyST, christopoulou-etal-2019-connecting, zhou2021document} by filtering negative relations like these, with disease entities that are hypernyms of a corresponding true relations disease entity within the same abstract, according to the hierarchy in the MeSH vocabulary.\footnote{\url{https://meshb.nlm.nih.gov}}

\section{Parsing the models output}
\label{appendix:parsing-model-output}

At test time, our model autoregressively generates an output, token-by-token, using beam search decoding (see \textsection\ref{model}). In order to extract the predicted relations from this output, we apply the following steps. First, predicted token ids are converted to a string. We use the \texttt{decode()}\footnote{\url{https://huggingface.co/docs/transformers/main_classes/tokenizer\#transformers.PreTrainedTokenizerBase.decode}} method of the HuggingFace Transformers tokenizer \citep{hf-transformers} to do this. For example, after calling \texttt{decode()} on the predicted token ids, this string might look like:\\

\noindent
{\small
\texttt{monoamine oxidase b ; maob @GENE@ parkinson's disease ; pd @DISEASE@ @GDA@}
}\\

\noindent We then use regular expressions to extract any relations from this string that match our linearization schema (see \textsection\ref{linearization}), which produces a dictionary of nested lists, keyed by relation class:\\

\noindent
{\small
\begin{verbatim}
{
  "GDA": [
    [
      [["monoamine oxidase b", "maob"], "GENE"],
      [["parkinson's disease", "pd"], "DISEASE"]
    ]
  ]
}
\end{verbatim}
}

\noindent Finally, we apply some normalization steps to the entity mentions. Namely, we strip leading and trailing white space characters, sort entity mentions lexicographically (as their order is not important), and remove duplicate mentions. Similarly, we remove duplicate relations. These steps are applied to both target and model output strings. The F1-score can then be computed by tallying true positives, false positives and false negatives.

\section{Relaxed entity matching}
\label{appendix:relaxed-entity-matching}

The aim of DocRE is to extract relations at the \textit{entity}-level. However, it is common to evaluate these methods with a \say{strict} matching criteria, where a predicted entity \(\mathcal{P}\) is considered correct if and only if all its \textit{mentions} exactly match a corresponding gold entities mentions, i.e. \(\mathcal{P} = \mathcal{G}\). This penalizes model predictions that miss even a single coreferent mention, but are otherwise correct. A relaxed criteria, proposed in prior work \citep{jain-etal-2020-scirex} considers \(\mathcal{P}\) to match \(\mathcal{G}\) if more than 50\% of \(\mathcal{P}\)'s mentions belong to \(\mathcal{G}\), that is

\[
    \frac{|\mathcal{P} \cap \mathcal{G}|}{|\mathcal{P}|} > 0.5
\]

\noindent In this paper, alongside the strict criteria, we report performance using this relaxed entity matching strategy, denoted \say{relaxed}.

\section{Hyperparameters}
\label{appendix:hyperparameters}

\begin{table*}[t]
\centering
\caption{Hyperparameter values used for each corpus. Hyperparameters values when using entity hinting, if they differ from the values used without entity hinting, are shown in parentheses. Tuned indicates whether or not the hyperparameters were tuned on the validation sets.}
\label{tab:a2}
\small
\resizebox{\textwidth}{!}{%
\begin{tabular}{@{}lccccc@{}}
\toprule
Hyperparameter                           & Tuned?                 & CDR       & GDA       & DGM       & DocRED \\ \midrule
Batch size                               & \color{MyGreen} \cmark & 4         & 4         & 4         & 4      \\
Training epochs                          & \color{MyGreen} \cmark & 130 (70)  & 30 (25)   & 30 (45)   & 50     \\
Encoder learning rate                    & \color{red} \xmark     & 2e-5      & 2e-5      & 2e-5      & 2e-5   \\
Encoder weight decay                     & \color{red} \xmark     & 0.01      & 0.01      & 0.01      & 0.01   \\
Encoder re-initialized top \(L\) layers  & \color{MyGreen} \cmark & 1         & 1 (2)     & 1         & 1      \\
Decoder learning rate & \color{MyGreen} \cmark & 1.21e-4 (1.13e-4) & 5e-4 (4e-4) & 8e-4 (1.5e-5) & 7.8e-5 \\
Decoder input dropout                    & \color{red} \xmark     & 0.1       & 0.1       & 0.1       & 0.1    \\
Decoder hidden-to-hidden weights dropout & \color{red} \xmark     & 0.5       & 0.5       & 0.5       & 0.5    \\
Target embedding size                    & \color{red} \xmark     & 256       & 256       & 256       & 256    \\
No. heads in multi-head cross-attention  & \color{red} \xmark     & 6         & 6         & 6         & 6      \\
Beam size                                & \color{MyGreen} \cmark & 3 (2)     & 4 (1)     & 3 (2)     & 8      \\
Length penalty                           & \color{MyGreen} \cmark & 1.4 (0.2) & 0.8 (1.0) & 0.2 (0.8) & 1.4    \\
Max decoding steps                       & \color{red} \xmark     & 128       & 96        & 96        & 400    \\ \bottomrule
\end{tabular}%
}
\end{table*}

In \autoref{tab:a2}, we list the hyperparameter values used during evaluation on each corpus, with and without entity hinting. Select hyperparameters were tuned using Optuna \citep{optuna_2019}. The tuning process selects the best hyperparameters according to the validation set micro F1-score using the TPE (Tree-structured Parzen Estimator) algorithm \citep{Bergstra2011AlgorithmsFH}.\footnote{\url{https://optuna.readthedocs.io/en/stable/reference/generated/optuna.samplers.TPESampler.html}} During tuning, we use greedy decoding (i.e. beam size of one). Once optimal hyperparameters are found, we tune the beam size (bs) and length penalty (\(\alpha\)) using a grid search over the values \(\text{bs}=\{2...10\}\), with a step size of \(1\), and \(\alpha=\{0.2...2.0\}\), with a step size of \(0.2\).

\section{Baselines}
\label{appendix:baselines}

This section contains detailed descriptions of all methods we compare to in this paper. 

\subsection{Pipeline-based methods}

These methods are pipeline-based, assuming the entities are provided as input. Many of them construct a document-level graph using dependency parsing, heuristics, or structured attention and then update node and edge representations using propagation.

\begin{itemize}
    \item \citet{christopoulou-etal-2019-connecting} propose EoG, an edge-orientated graph neural model. The nodes of the graph are constructed from mentions, entities, and sentences. Edges between nodes are initially constructed using heuristics. An iterative algorithm is then used to generate edges between nodes in the graph. Finally, a classification layer takes the representation of entity-to-entity edges as input to determine whether those entities express a relation or not. We compare to EoG in the pipeline-based setting on the CDR and GDA corpora.
    \item \citet{nan-etal-2020-reasoning} propose LSR (Latent Structure Refinement). A \say{node constructor} encodes each sentence of an input document and outputs contextual representations. Representations that correspond to mentions and tokens on the shortest dependency path in a sentence are extracted as nodes. A \say{dynamic reasoner} is then applied to induce a document-level graph based on the extracted nodes. The classifier uses the final representations of nodes for relation classification. We compare to LSR in the pipeline-based setting on the CDR and GDA corpora.
    \item \citet{lai2020bert} propose BERT-GT, which combines BERT with a graph transformer. Both BERT and the graph transformer accept the document text as input, but the graph transformer requires the neighbouring positions for each token, and the self-attention mechanism is replaced with a neighbour–attention mechanism. The hidden states of the two transformers are aggregated before classification. We compare to BERT-GT in the pipeline-based setting on the CDR and GDA corpora.
    \item \citet{minh-tran-etal-2020-dots} propose EoGANE (EoG model Augmented with Node Representations), which extends the edge-orientated model proposed by \citet{christopoulou-etal-2019-connecting} to include explicit node representations which are used during relation classification. We compare to EoGANE in the pipeline-based setting on the CDR and GDA corpora.
    \item SSAN \citep{Xu2021EntitySW} propose SSAN (Structured Self-Attention Network), which inherits the architecture of the transformer encoder \citep{vaswani2017attention} but adds a novel structured self-attention mechanism to model the coreference and co-occurrence dependencies between an entities mentions. We compare to SSAN in the pipeline-based setting on the CDR and GDA corpora.
    \item \citet{zhou2021document} propose ALTOP (Adaptive Thresholding and Localized cOntext Pooling), which extends BERT with two modifications. Adaptive thresholding, which learns an optimal threshold to apply to the relation classifier. Localized context pooling, which uses the pretrained self-attention layers of BERT to create an entity embedding from its mentions and their context. We compare to ALTOP in the pipeline-based setting on the CDR and GDA corpora.
\end{itemize}

\subsection[n-ary relation extraction]{\(n\)-ary relation extraction}

These methods are explicitly designed for the extraction of \(n\)-ary relations, where \(n > 2\). 

\begin{itemize}
    \item \citet{jia-etal-2019-document} propose a multiscale neural architecture, which combines representations learned over text spans of varying scales and for various sub-relations. We compare to \citet{jia-etal-2019-document} in the pipeline-based setting on the \(n\)-ary DGM corpus.
\end{itemize}

\subsection{End-to-end methods}

These methods are capable of performing the subtasks of DocRE in an end-to-end fashion with only the document text as input.

\begin{itemize}
    \item \citet{eberts-ulges-2021-end} propose JEREX, which extends BERT with four task-specific components that use BERTs outputs to perform entity mention localization, coreference resolution, entity classification, and relation classification. They present two versions of their relation classifier, denoted \say{global relation classifier} (GRC) and \say{multi-instance relation classifier} (MRC). We compare to JEREX-MRC in the end-to-end setting on the DocRED corpus.
\end{itemize}

\end{document}